\documentclass{article}

\usepackage[preprint]{neurips_2026}

\usepackage[utf8]{inputenc} % allow utf-8 input
\usepackage[T1]{fontenc}    % use 8-bit T1 fonts
\usepackage{hyperref}       % hyperlinks
\usepackage{url}            % simple URL typesetting
\usepackage{booktabs}       % professional-quality tables
\usepackage{amsfonts}       % blackboard math symbols
\usepackage{nicefrac}       % compact symbols for 1/2, etc.
\usepackage{microtype}      % microtypography
\usepackage{xcolor}         % colors
\usepackage{graphicx}
\usepackage{subcaption}
\usepackage{float}
\usepackage{amsmath}
\hypersetup{colorlinks=true, linkcolor=black, citecolor=black, urlcolor=blue}

\title{Learning Compression Rules for Network Traffic}

\author{
    Quentin Lampin$^{1}$ \And Éloi Sainte-Beuve$^{1,2}$ \AND Louis-Adrien Dufrène$^{1}$ \And Guillaume Larue$^{1}$ \And Massih-Reza Amini$^{2}$
    \AND $^{1}$Orange Research \quad $^{2}$Université Grenoble Alpes \\
    \texttt{\{quentin.lampin, eloi.sainte-beuve\}@orange.com} \\
    \texttt{\{louisadrien.dufrene, guillaume.larue\}@orange.com} \\
    \texttt{massih-reza.amini@univ-grenoble-alpes.fr}
}

\begin{document}

\maketitle

\begin{abstract}
    We study the problem of learning compact rule-based compressors for
    structured network traffic. Each packet is a record of header fields
    that are highly redundant within a flow, and a compressor is a small
    set of rules matching such records and replacing predictable fields
    with short codes. We cast rule learning as a two-stage problem:
    (i) an unsupervised structure-discovery stage that recursively
    partitions training packets using a normalized entropy-ratio
    criterion robust to small samples, and (ii) a constrained selection
    stage that uses dynamic programming to pick the rule subset
    maximizing expected compression gain under a hard budget on the
    number of installable rules. We instantiate the framework on Static
    Context Header Compression (SCHC), the IETF standard for rule-based
    header compression in constrained networks, and evaluate it on four
    real-world Internet-of-Things and 5G core-network datasets. Our
    method, Robust Entropy Clustering for Adaptive comPression (RECAP),
    surpasses expert-engineered rule sets with a small number of learned
    rules and removes the need for manual rule design.
\end{abstract}

\section{Introduction}
\label{sec:introduction}

We study how to learn small rule-based compressors directly from samples of
structured network traffic. A network packet can be seen as a fixed-schema
record whose fields---addresses, ports, protocol identifiers, sequence
numbers, and so on---carry both useful information and a large amount of
metadata that is highly redundant within a given flow: many fields stay
constant or vary in a predictable way across consecutive packets. Header
compression methods exploit this redundancy by replacing such fields with
short residues, using a context of rules shared between sender and receiver.
The quality of these rules is what determines the achievable compression
ratio.

In practice, such rules are written by hand. The Static Context Header
Compression (SCHC) standard~\cite{minaburo_schc_2020}, defined by the
Internet Engineering Task Force (IETF) for Low-Power Wide-Area Networks
(LPWANs) and now extended to other settings such as Ethernet and tunneling
protocols~\cite{pelov_static_2023,thubert_schc_2023}, specifies how rules
should be encoded and applied at runtime, but it does not say how to obtain
them. Companion Request for Comments (RFC) profiles recommend rule sets for
common Internet-of-Things (IoT) protocols~\cite{minaburo_static_2021,
gimenez_static_2021,zuniga_static_2023,ramos_static_2023}, but constructing
an efficient rule set still requires combining detailed knowledge of every
protocol layer with manual analysis of the targeted traffic. This expertise
rarely sits in a single team, and the resulting rules are hard to maintain
across new applications, deployments, or protocol versions.

We propose to replace this manual loop by data-driven learning. Given a
packet capture, we treat each parsed packet as a record of header fields
and ask: which subset of rules, drawn from the structure of the data,
maximizes the expected compression gain under a fixed rule budget? Our
method, Robust Entropy Clustering for Adaptive comPression (RECAP),
answers this question in two stages:

\begin{itemize}
    \item \textbf{Structure discovery.} We recursively partition the
        training packets to expose the natural clusters of the traffic.
        At each step we select one field whose conditional values
        best regularize the cluster, using a normalized entropy ratio designed 
        to remain reliable when the field's empirical support is sparse relative
        to its true alphabet size, and is paired with a finite-sample normalized-entropy 
        stopping rule. This yields a partition tree of candidate clusters.
    \item \textbf{Constrained rule selection.} Each node of the resulting
        partition tree yields a candidate rule with an empirically estimated
        expected compression gain. We then use dynamic programming to
        pick the subset of candidates that maximizes the total expected
        gain subject to a hard cap on the number of installable rules.
\end{itemize}

We instantiate the framework on SCHC because it is the most widely
standardized rule-based header compressor, but the underlying output ---
a data-driven inventory of constant, low-cardinality, and
high-entropy fields in a given trace, together with a budgeted selection
of templates that exploit them --- can prime any compressor that benefits from
such priors: it yields rule sets directly for static rule-based schemes, and
provides initialization material for stateful frameworks such as
ROHC~\cite{jonsson_robust_2010}.
We evaluate RECAP on four real-world datasets spanning IoT and 5G core-network 
traffic, and show that a small number of learned rules already match expert-engineered
baselines.

The remainder of the paper is organized as follows. Section
\ref{sec:schc-explained} introduces just enough of SCHC to follow the rest
of the paper. Section \ref{sec:related-work} positions our contribution
with respect to prior work. Section \ref{sec:detailed-proposal} presents
the RECAP algorithm, and Sections \ref{sec:performance-eval}, \ref{sec:limitations} 
report experimental results and limitations. Section \ref{sec:conclusion} concludes.

\section{SCHC Compression/Decompression}
\label{sec:schc-explained}

\begin{figure}[htbp]
    \centering
    \begin{minipage}[t]{0.45\linewidth}
        \vspace{0pt}
        \centering
        \includegraphics[width=\linewidth,trim=0 24 0 18,clip]{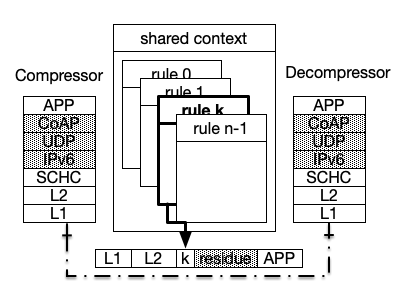}
        \caption{SCHC illustration: rule $k$ from the shared context is used to
        compress the CoAP, UDP and IPv6 headers of a packet.}
        \label{fig:schc-context}
    \end{minipage}\hfill
    \begin{minipage}[t]{0.52\linewidth}
        \vspace{0pt}
        \captionsetup{type=table}
        \centering
        \scriptsize
        \setlength{\tabcolsep}{3pt}
        \resizebox{\linewidth}{!}{%
        \begin{tabular}{|l|c|c|c|c|c|}
                \hline
                \multicolumn{6}{|c|}{\textbf{Compression Rule ID: 1}} \\
                \hline
                \textbf{FID}             & \textbf{LEN} & \textbf{POS} & \textbf{DI} & \textbf{MO/CDA}   & \textbf{TV} \\
                \hline
                IPv6:Version             & 4            & 0            & Bi          & Equal/Not-Sent    & [06] \\
                IPv6:Traffic Class       & 8            & 0            & Bi          & Equal/Not-Sent    & [00] \\
                IPv6:Flow Label          & 20           & 0            & Up          & Ignore/Value-Sent & N/A \\
                IPv6:Payload Length      & 16           & 0            & Up          & MSB/LSB           & [00 00](10) \\
                IPv6:Next Header         & 8            & 0            & Up          & Equal/Not-Sent    & [11] \\
                IPv6:Hop Limit           & 8            & 0            & Up          & Equal/Not-Sent    & [40] \\
                IPv6:Source Address      & 128          & 0            & Up          & Equal/Not-Sent    & [20010db8::03] \\
                IPv6:Destination Address & 128          & 0            & Up          & Equal/Not-Sent    & [20010db8::20] \\
                UDP:Source Port          & 16           & 0            & Up          & Equal/Not-Sent    & [c9ad] \\
                UDP:Destination Port     & 16           & 0            & Up          & Equal/Not-Sent    & [1633] \\
                UDP:Length               & 16           & 0            & Up          & Ignore/Compute    & N/A \\
                UDP:Checksum             & 16           & 0            & Up          & Ignore/Compute    & N/A \\
                \multicolumn{6}{|c|}{\dots} \\
                CoAP:Payload Marker      & 8            & 0            & Bi          & Equal/Not-Sent    & [ff] \\
                \hline
            \end{tabular}}
            \vspace{6pt}
        \caption{Compact excerpt of a compression rule for a CoAP request.}
        \label{table:compression-rule}
    \end{minipage}
\end{figure}

SCHC~\cite{minaburo_schc_2020} is a rule-based header compressor: sender
and receiver share a static \emph{context} of compression/decompression
rules, and each rule describes how to compress one particular kind of
packet. At runtime, compression replaces the matched header fields with a
compact residue prefixed by the rule identifier; decompression reverses
the operation by reading the rule identifier and reconstructing each field
from the rule. Figure~\ref{fig:schc-context} illustrates the shared-context
mechanism on a packet whose header layers are Internet Protocol version~6
(IPv6), User Datagram Protocol (UDP), and Constrained Application Protocol
(CoAP)---the three layers that appear in many IoT deployments.

A rule is a small table with one row per header field; an excerpt for a
CoAP request is shown in Table~\ref{table:compression-rule}. Each row
identifies a field by its protocol-qualified name (FID, e.g.
\texttt{IPv6:Source\,Address}) together with its bit-length (LEN) and
position (POS) inside the header (when the field is present multiple times), 
and the traffic direction (DI) it applies to (uplink, downlink, or bidirectional).
The remaining columns specify how the field is handled at runtime: a matching 
operator together with a compression/decompression action (MO/CDA) compares the 
observed value to a target value (TV) and decides whether the field is omitted,
sent verbatim, sent as a least-significant-bits residue, or replaced by
an index in a small mapping. A few additional actions cover fields that
can be recomputed at the receiver (e.g., UDP length and checksum) or
rebuilt from link-layer addresses; we refer to RFC~8724 for the full
operator set.

A packet \emph{matches} a rule when every row's matching operator accepts
the corresponding observed value, in which case the rule is applied row
by row to produce the compressed residue. Decompression performs the
inverse actions to rebuild the original header. For the rest of the
paper, the only thing to remember is that a SCHC rule is a structured
template that simultaneously describes \emph{which packets it accepts}
and \emph{how their fields are compressed}, so learning a good rule set
amounts to discovering useful templates from data and selecting a small
subset of them.

\section{Related Work}
\label{sec:related-work}

Most SCHC literature focuses on standards definition, protocol adaptations, and
deployment-specific evaluation rather than automated rule learning. Foundational
specifications define SCHC as a static-context mechanism where rule quality is
critical but generally engineered manually \cite{minaburo_schc_2020,pelov_static_2023}.
Empirical studies of real-world SCHC deployments confirm this practice and
report performance for expert-designed rule
sets~\cite{sisinni_performance_2023,dumay_effective_2021}.
Follow-up RFCs and profiles extend SCHC to specific stacks and link layers
(CoAP, LoRaWAN, Sigfox, NB-IoT, PPP), but still assume expert-crafted rule sets
\cite{minaburo_static_2021,gimenez_static_2021,zuniga_static_2023,ramos_static_2023,thubert_schc_2023}.
Two recent works move toward learning-based automation. Banerjee et
al.~\cite{banerjee_automated_2024} frame SCHC rule generation as a
pattern-recognition problem using flat clustering on Gower field-by-field
distances and a heuristic to construct one SCHC rule per cluster. We adapted
this approach to our experimental framework to serve as a direct clustering
baseline (Appendix~\ref{appendix:ablations}); it performs competitively on
homogeneous IoT traces but collapses on heterogeneous 5G traffic where flat,
structure-agnostic grouping fragments the budget across too many small clusters.
Meslet-Millet et al.~\cite{meslet-millet_dch_2023}
take an orthogonal route with DCH, an end-to-end deep-learning header codec that
compresses arbitrary header bytes through a neural network shared between sender
and receiver. DCH bypasses the rule abstraction entirely: it does not produce 
SCHC-compatible rules, requires neural inference at both endpoints, and
therefore cannot be deployed inside the IETF SCHC context. RECAP differs from 
both: it learns \emph{interpretable, standards-compliant} SCHC rules, runs no 
model at runtime, and integrates directly into any RFC~8724 endpoint.

Beyond SCHC, the broader header-compression ecosystem includes
Robust Header Compression (ROHC), 6LoWPAN-GHC,
and end-to-end forwarding frameworks \cite{jonsson_robust_2010,bormann_6lowpan-ghc_2014,jia_end--end_2022}.
The common trend is still hand-designed profiles and heuristics, with learning
used mainly for tuning or architecture-level decisions. A comprehensive survey by
Tömösközi et al. \cite{tomoskozi_packet_2022} confirms this gap: despite strong
progress in protocol engineering, principled data-driven rule discovery remains
underdeveloped, especially when optimization must account jointly for statistical
coverage and a strict rule budget.

For readers seeking broader context outside networking headers, general-purpose
lossless compression is thoroughly covered in survey-style references and
handbooks \cite{salomon_handbook_2010,sayood_introduction_2017}. These works
review classical coding and dictionary-based families (e.g., Huffman, arithmetic
coding, LZ variants) and formalize entropy limits that motivate our design.

\section{Robust Entropy Clustering for Adaptive comPression (RECAP)}
\label{sec:detailed-proposal}
Our method, \textbf{Robust Entropy Clustering for Adaptive comPression (RECAP)},
specializes information-theoretic algorithms to SCHC rule synthesis. RECAP has 
two phases: \textbf{Packet Partitioning} (divisive entropy-guided clustering for
candidate generation) followed by \textbf{Rule Selection} (dynamic programming
for selection under a fixed rule budget).

\subsection{Setup}
\label{sec:recap-setup}

For a fixed protocol structure, we represent a packet as a random vector
$X=(X_1,\dots,X_d)$ where each header field $X_j$ takes values in an
alphabet $\mathcal{X}_j$. Given $n$ training packets
$x^{(1)},\dots,x^{(n)}$, we write $x_A^{(i)}$ for the restriction of
packet $x^{(i)}$ to the fields in $A$. The empirical joint distribution
of a field subset $A\subseteq\{1,\dots,d\}$ is
\begin{equation}
\hat{p}_A(v_A)=\frac{1}{n}\sum_{i=1}^n \mathbf{1}\{x_A^{(i)}=v_A\},
\qquad
\hat{H}(X_A)=-\!\!\sum_{v_A\in\hat{\mathcal{X}}_A}\!\!
\hat{p}_A(v_A)\log_2\hat{p}_A(v_A),
\end{equation}
where $\hat{\mathcal{X}}_A$ is the observed support and $\hat{H}$ denotes
the plug-in entropy estimator. Two regime properties shape the rest of
the design. First, header alphabets are very large (an IPv6 address alone
covers $2^{128}$ values), so any cluster's training sample is a sparse
draw from the true distribution and rules must generalize to unseen
values. Second, $\hat{H}$ is negatively biased on small samples,
$\mathbb{E}[\hat{H}(X_A)]\le H(X_A)$, with the bias most severe exactly
when $n$ is small relative to $|\mathcal{X}_A|$. Both effects motivate
the normalized splitting criterion and the normalized-entropy stopping rule
below.

\subsection{Packet Partitioning via Divisive Clustering}

We build a hierarchy of candidate clusters by recursively partitioning
the training set. The root cluster contains every training packet. We first
perform a \emph{structural} split: because SCHC matches each field by
identifier, length, and position, packets with different header layouts
cannot share a rule, so the root cluster is partitioned into one child per
observed sequence of field descriptors. We then apply \emph{value-based}
splits recursively to each structurally homogeneous child: we select one
field index $j\in\{1,\dots,d\}$ that best regularizes the cluster (criterion
below), and create one grandchild per distinct value of $X_j$.
Each child cluster is defined by a single conditioning value $X_j = v$: by
construction, every packet in that child shares the same value for field $j$,
so field $j$ is constant within the child and can be elided by a SCHC rule
with \texttt{MO:Equal} / \texttt{CDA:Not-Sent}.
Splits accumulate across levels: for instance, a node may first split on
CoAP type, and a child node may then split on URI path, making both fields
constant in the resulting grandchild.

Selecting $j$ from $\hat{H}(X_j)$ alone is unreliable: the empirical
estimator is bounded by $\hat{H}(X_j)\le\log_2(n_c)$ on a cluster of
size $n_c$, which suppresses the apparent entropy of large-alphabet
fields when $n_c$ is small, and raw entropy ignores how many bits a
field actually occupies. We therefore select the splitting field as
\begin{equation}
    j^* = \arg\min_{j\,:\,\kappa_j \ge 2}\, R(j),
    \qquad
    R(j) =
    \begin{cases}
        \dfrac{\hat{H}(X_j)}{\min(\hat{L}_j,\,\log_2(n_c))} & \text{if } n_c > 1, \\[4pt]
        0 & \text{if } n_c = 1,
    \end{cases}
\end{equation}
where $\hat{L}_j$ is the empirical mean bit-length of field $j$ in the
cluster and $\kappa_j$ denotes the number of distinct values taken by
$X_j$ in the cluster; the minimum is restricted to non-constant fields
($\kappa_j{\ge}2$), since constant fields are already fully compressible. 
The denominator balances compressibility with statistical confidence: when
$\log_2(n_c)\ge\hat{L}_j$ the ratio normalizes by field width and rewards
density, while when the sample is sparse ($\log_2(n_c)<\hat{L}_j$) it
saturates at $\log_2(n_c)$ and prevents selecting a field whose low
apparent entropy is merely a small-sample artifact. We split the cluster
if $R(j^*) < \theta$, and stop otherwise; the
recursion also terminates when the cluster is pure (all packets share
identical field values).

This threshold has two complementary interpretations. When
$\log_2(n_c)\ge\hat{L}_j$, the denominator is $\hat{L}_j$, so
$R(j)=\hat{H}(X_j)/\hat{L}_j$ is the fraction of the field bit-length that is
effectively informative (non-redundant): a high ratio means little practical
compressibility remains, so further splitting is not useful. When
$\log_2(n_c)<\hat{L}_j$, the denominator becomes $\log_2(n_c)$: in this
sparse-sample regime, the threshold acts as a statistical guardrail that limits
splits on fields for which the available observations provide weak support.
The result is a tree where each node represents a subset of packets (a
cluster) definable by a specific set of field constraints. Figure~\ref{fig:recap-flow}
illustrates the resulting hierarchy on an IoT traffic capture: the first level
reflects the structural pre-split into one child per observed field-descriptor
sequence, and subsequent levels are value-based splits driven by the entropy-ratio
criterion.

\subsection{SCHC Rule Generation}
\label{sec:schc-rule-generation}
Each cluster in the hierarchy yields one candidate rule. For every field
$X_j$, RECAP assigns a matching operator and compression/decompression action
(MO/CDA) based on the empirical distribution of values observed in the cluster
and the entropy-ratio statistic. Let $M_{\mathrm{map}}$ be a hyperparameter
capping the size of a per-field mapping table,
and recall the field cardinality $\kappa_j$ and the normalized entropy
ratio $R(j)$ from Section~\ref{sec:recap-setup}. The assignment heuristic is:

\begin{table}[htbp]
    \centering
    \footnotesize
    \setlength{\tabcolsep}{5pt}
    \begin{tabular}{p{0.42\linewidth}p{0.3\linewidth}p{0.25\linewidth}}
        \toprule
        \textbf{Condition} & \textbf{MO/CDA} & \textbf{Transmitted} \\
        \midrule
        Field is recomputable & Ignore / Compute & Nothing \\
        $\kappa_j = 1$ & Equal / Not-Sent & Nothing \\
        $1 < \kappa_j \le M_{\mathrm{map}}$ and $R(j) < \theta$ & Match-Mapping / Mapping-Sent & Index ($\lceil\log_2\kappa_j\rceil$ bits) \\
        Otherwise & Ignore / Value-Sent & Full field value \\
        \bottomrule
    \end{tabular}
\end{table}

\noindent
Recomputable fields are always marked for computation (Ignore/Compute) regardless
of cardinality or entropy, since they can be deterministically reconstructed.
Constant fields ($\kappa_j=1$) are always marked for elision (Equal/Not-Sent)
while fields with low cardinality ($1 < \kappa_j \le M_{\mathrm{map}}$) and low
entropy ratio ($R(j) < \theta$) are marked for mapping (Match-Mapping/Mapping-Sent), 
which replaces the field value with a compact index in a small mapping table.
The entropy-ratio gating ($R(j) < \theta$) in the third row reflects
the threshold's two interpretations. When sample size is adequate, $R(j)$
measures the fraction of field bit-length that is effectively informative:
a high ratio signals little practical compressibility, so compression is avoided.
When the sample is sparse, $R(j)$ acts as a statistical guardrail: it suppresses
compression of fields for which the observed homogeneity may be a sampling
artifact rather than a true structural property. All other fields---whether
they fail the entropy threshold, have high cardinality, or are neither
predictable nor recomputable---are sent in full (Ignore/Value-Sent). These
rules serve as \emph{candidate rules} for the subsequent rule-selection phase.

\subsection{Rule Selection via Dynamic Programming}
\label{sec:rule-selection}
\subsubsection{Setup and Coverage Estimation}

The candidate hierarchy is a tree rooted at a special \emph{no-compression rule}
that represents all training packets without any field compression. This root node
accommodates packets with different structural compositions (different sequences of
header fields). The root's immediate children partition the packets by structure,
so each subtree rooted at a child contains structurally homogeneous packets that
can share a common SCHC rule. We write $V$ for the set of nodes in this
candidate tree; for each node $u\in V$ we build a candidate SCHC rule that
compresses its packets according to the field heuristic in
Section~\ref{sec:schc-rule-generation}.

To estimate whether a rule will apply to future packets (not in the training set),
we use the \emph{coverage} of the rule: the probability that an incoming packet's
field values match the rule's value constraints. Let $\mathcal{T}_u$ be the
multiset of joint value tuples for the fields whose values are constrained by the
rule built from cluster $u$ --- that is, fields assigned Equal/Not-Sent or
Match-Mapping/Mapping-Sent --- across the $n_u$ training packets of cluster $u$.
Let $f_1(u)$ be the number of distinct tuples in $\mathcal{T}_u$ that appear
exactly once (singletons). The Good-Turing sample coverage estimator~\cite{good_population_1953} is:
\begin{equation}
\hat{C}(u) = 1 - \frac{f_1(u)}{n_u}.
\end{equation}
The estimator subtracts the singleton mass from $1$, yielding the
probability that a new packet's tuple has already been observed at
least twice in training. Future packets whose tuples were seen exactly
once may still legitimately match the rule, so $\hat{C}(u)$ is a
conservative (lower-bound) proxy for the true rule-firing probability
under i.i.d.\ future traffic; this conservatism biases the DP toward
rules with stable, repeatedly-observed constraints. A large $f_1(u)/n_u$
indicates a heterogeneous cluster with many one-off patterns; a small ratio
indicates a stable, generalizable rule.

\subsubsection{Dynamic Programming Formulation}

\paragraph{Problem.} Out of all candidate rules in the tree, we must pick a
subset of size at most $N$ that maximizes total expected compression gain.
A naive enumeration over $\binom{|V|}{N}$ subsets is impractical, but the
tree structure constrains the choice: rules are organized hierarchically,
and at every node $u$ we face the same elementary decision --- \emph{select}
$u$ or \emph{skip} it. This local structure turns the global selection
problem into a budget-allocation problem between each node and its children,
which we solve by tree dynamic programming~\cite{kellerer_knapsack_2004}.

\paragraph{Local choice and budget allocation.} Consider a node $u$ with a
budget of $k$ rules to spend in its subtree, given that the closest already
selected ancestor is $a$ (initially the no-compression root). Two scenarios
arise:
\begin{itemize}
    \item \textbf{Select $u$:} consume one credit, install the candidate
        rule for $u$, and distribute the remaining $k-1$ credits among
        $u$'s children. Because $u$ is now installed, it replaces $a$ as
        the active ancestor for its children.
    \item \textbf{Skip $u$:} install no rule at $u$, keep $a$ as the active
        ancestor, and distribute the full budget $k$ among $u$'s children.
\end{itemize}
Choosing the better of the two scenarios at every node, recursively, yields
the optimum subset of size $\leq N$ \emph{within the candidate set}
\footnote{we do not claim optimality over the space of all SCHC rule sets, since
the preceding divisive clustering is greedy.}.

\paragraph{Scoring a selection.} To compare the two scenarios we need the
contribution of installing rule $u$ when $a$ is the active ancestor. We
define the \emph{expected incremental gain}
\begin{equation}
g(u \mid a) = \hat{C}(u) \cdot \bigl(\gamma(u) - \gamma(a \to u)\bigr),
\end{equation}
where $\gamma(u)$ is the total compression gain obtained by applying rule
$u$ to each of the $n_u$ packets of cluster $u$, $\gamma(a \to u)$ is the
corresponding gain obtained by applying ancestor rule $a$ to those same
packets, and $\hat{C}(u)$ is the Good-Turing coverage from the previous
subsection.
The difference $\gamma(u) - \gamma(a \to u)$ isolates the marginal benefit
of selecting $u$ on top of $a$: rule $u$ is only credited for what it adds
beyond the fallback ancestor. Multiplying by $\hat{C}(u)$ down-weights
rules whose value constraints are unlikely to fire on unseen packets, so
the score reflects expected behavior on future traffic rather than raw
fit to the training set.

\paragraph{Recurrence.} Let $G(u, k \mid a)$ denote the maximum expected
gain achievable in the subtree rooted at $u$ with at most $k$ rules,
given active ancestor $a$. With children
$\mathrm{ch}(u) = \{v_1, \dots, v_m\}$, the two scenarios above give:
\begin{equation}
\begin{split}
G(u, k \mid a) = \max\Bigl\{\;
  &\max_{k_1+\cdots+k_m=k}\;
    \sum_{v\in\mathrm{ch}(u)} G(v, k_v \mid a),
    \hfill\text{[skip } u\text{]} \\[4pt]
  &g(u \mid a) +
    \max_{k_1+\cdots+k_m=k-1}\;
    \sum_{v\in\mathrm{ch}(u)} G(v, k_v \mid u)
  \;\Bigr\}, \hfill\text{[select } u\text{]}
\end{split}
\end{equation}
with base cases $G(u, 0 \mid a) = 0$ and
$G(\mathrm{leaf}, k \mid a) = \max\bigl(0,\, g(\mathrm{leaf} \mid a)\bigr)$
for $k \ge 1$ (a leaf may be skipped if the marginal gain is negative).
Note that the active-ancestor argument flips from $a$ to $u$ in the
``select'' branch: subsequent decisions in the subtree are taken \emph{relative
to $u$}, not to $a$.

\begin{figure}[htbp]
    \centering
    \includegraphics[width=\linewidth]{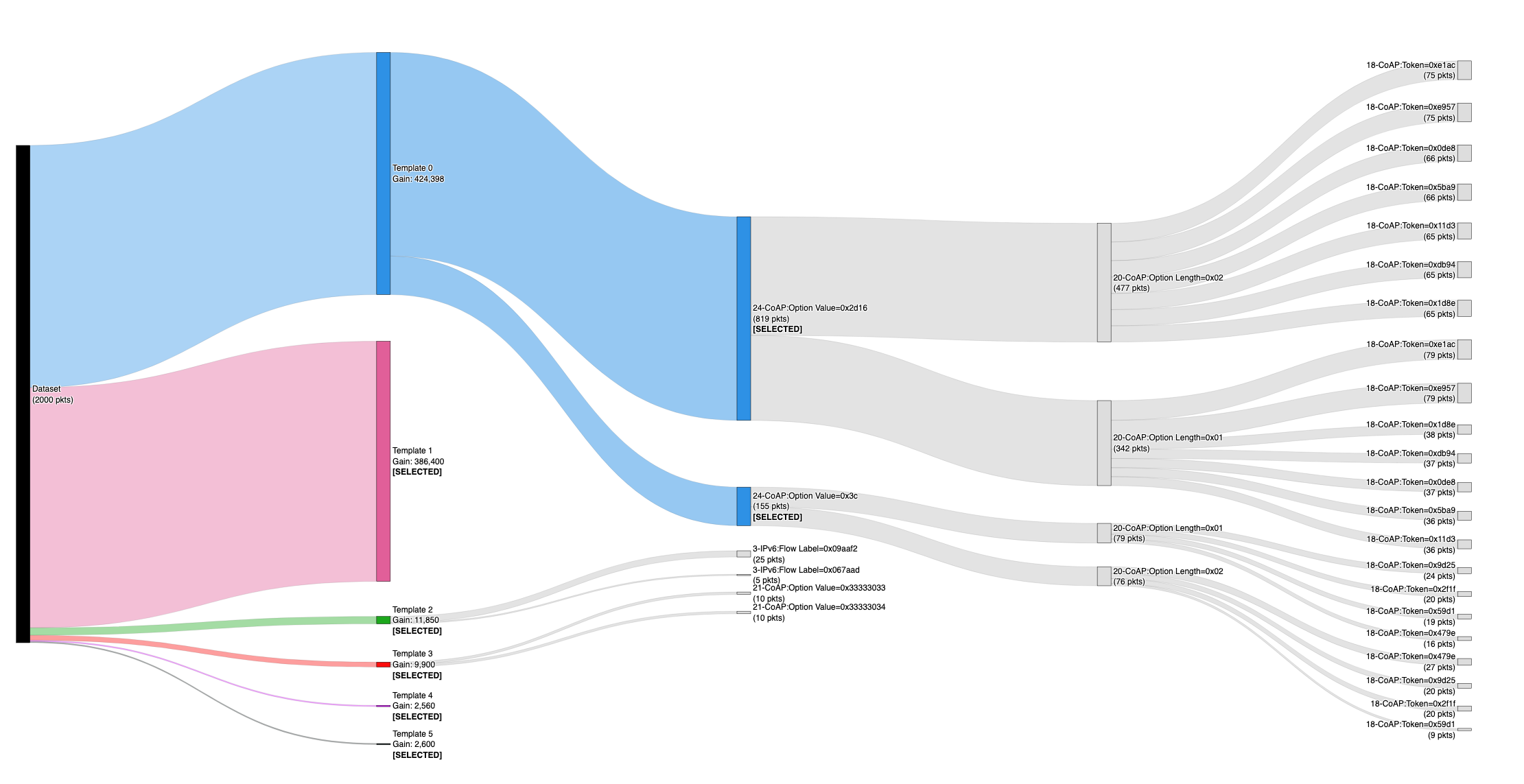}
    \caption{RECAP candidate hierarchy on an IoT traffic capture
    (Balloon-20k, 10\% training split). The root
    contains all training packets; its first-level children correspond
    to the structural pre-split, one per observed sequence of field
    descriptors, with each layout shown in a distinct color. Subsequent
    levels are per-value splits driven by the entropy-ratio criterion.
    Highlighted edges trace the $8$ rules selected by the dynamic
    program at $N=8$; greyed edges are explored but unselected
    candidates.}
    \label{fig:recap-flow}
\end{figure}

\paragraph{Objective.} The rule set under budget $N$ is recovered
from
\begin{equation}
G^\star = G(\mathrm{root},\, N \mid \emptyset),
\end{equation}
where $\emptyset$ denotes the absence of any prior rule. The inner budget-allocation 
step over $m$ children is a tree-knapsack subproblem solved in $O(N^2)$ per node
by standard convolution, giving overall complexity $O(|V|\cdot N^2)$. Highlighted 
edges in Figure~\ref{fig:recap-flow} trace the $N=8$ rules selected by the DP; 
greyed nodes are evaluated but not selected.

% \paragraph{Scope of the guarantee.} The recurrence returns the subset of
% size at most $N$ that maximizes expected gain \emph{within the candidate
% hierarchy} produced in Section~\ref{sec:detailed-proposal}. Because the
% partitioning step is greedy --- one field selected per split via $R(j)$
% --- it does not enumerate all conceivable SCHC rule sets; the DP is
% exact relative to the candidates it is given, not relative to the
% unrestricted rule-set space. Decoupling candidate generation from
% selection is what makes the budgeted problem tractable: the candidate
% tree compresses an otherwise super-exponential search space into a
% $|V|$-node structure on which budgeted optimization runs in polynomial
% time.

\subsubsection{Computational Optimizations}

Two reductions tighten the search space without sacrificing optimality.
\textbf{Budget bounding} caps $N_u \le |V_u|$ since at most $|V_u|$ rules
are selectable in subtree $u$. \textbf{Branch-and-bound pruning} uses
$\gamma(u)$ (the absolute compression gain assuming no ancestor) as an
admissible upper bound on $g(u \mid a)$ for any ancestor $a$, allowing
early termination when the remaining budget cannot improve the best
solution found so far.

\section{Performance Evaluation}
\label{sec:performance-eval}

\subsection{Experimental Setup}

We evaluate RECAP on four real-world traces spanning two distinct contexts:
two IoT/CoAP captures (\textbf{Balloon-20k}, \textbf{Thermostat-10k}) on
which SCHC was originally designed, and two cellular core-network captures
(\textbf{GTP-traffic}, \textbf{NGAP-traffic}) that probe regimes outside
SCHC's classical scope --- one a $100$-packet small-sample 2G/GPRS trace,
the other a heterogeneous $15{,}650$-packet 5G N2 control-plane capture
with deep ASN.1-encoded signaling structures.
Table~\ref{table:datasets} summarizes their basic statistics; per-dataset
qualitative descriptions and their structure recovered
on each trace are deferred to Appendix~\ref{appendix:templates}.
For GTP and NGAP, parsers expose protocol fields all the way down to
information elements rather than separating a fixed application payload, so
the parsed packet representation is treated entirely as header content and
the header ratio is therefore $100\%$.

\begin{table}[htbp]
    \centering
    \begin{tabular}{llll}
        \toprule
        \textbf{Dataset} & \textbf{Packets} & \textbf{Protocol Stack} &
            \textbf{Header Ratio} \\
        \midrule
        Balloon-20k    & $20{,}000$ & IPv6/UDP/CoAP   & $92.9\%$ \\
        Thermostat-10k & $10{,}000$ & IPv6/UDP/CoAP   & $86.8\%$ \\
        GTP-traffic    & $100$      & IPv4/UDP/GTPv1  & $100.0\%$ \\
        NGAP-traffic   & $15{,}650$ & IPv4/SCTP/NGAP  & $100.0\%$ \\
        \bottomrule
    \end{tabular}
    \caption{Dataset summary. The header ratio
    $(1 - \text{payload\_bits}/\text{total\_bits}) \times 100\%$ is an upper
    bound on compression (all header bits eliminated).}
    \label{table:datasets}
\end{table}

For each dataset we sweep a full grid of $N \in
\{2, 3, 4, 5, 6, 7, 8, 10, 12, 14, 16, 24, 32\}$ rules and four training-split
ratios $\{10\%, 20\%, 40\%, 50\%\}$. 

Splits are deterministic chronological prefixes: the first $r\%$ of packets in
capture order are used for training and the remaining $(1-r)\%$ form the 
held-out test set on which all reported numbers are computed. This reflects the
operational regime in which a SCHC rule set is built --- only past traffic is
available at deployment time --- and prevents the leakage that uniform random
shuffling would silently introduce. 

The normalized-entropy stopping threshold is
fixed at $\theta = 0.95$ and the mapping-table cap at $M_{\mathrm{map}}=8$
entries throughout.

Rule sets are encoded and evaluated using an open-source Python
implementation of RFC~8724~\cite{microschc}; recomputable fields
(IPv6/UDP length and checksums, SCTP checksum) are marked as such and all
others are assigned MO/CDA by the entropy-ratio heuristic of
Section~\ref{sec:schc-rule-generation}. 

At runtime, every admissible rule
is evaluated against the packet and the one yielding the shortest output
is used, so each packet is compressed by the most specific applicable
rule.

For all four datasets we additionally report a simple expert-style
baseline: the RFC~8824~\cite{minaburo_static_2021} SCHC profile on the
IoT datasets, and a structurally-derived rule set on the 5G datasets,
built from the published RFC/3GPP transport-header structure only and
treating bytes past those headers as opaque payload (construction details
in Appendix~\ref{appendix:templates}).

The full experimentation campaign runs on a single laptop in a few hours.

\subsection{Main Results}

\begin{figure}[htbp]
    \centering
    \begin{subfigure}[t]{0.49\linewidth}
        \centering
        \includegraphics[width=\linewidth]{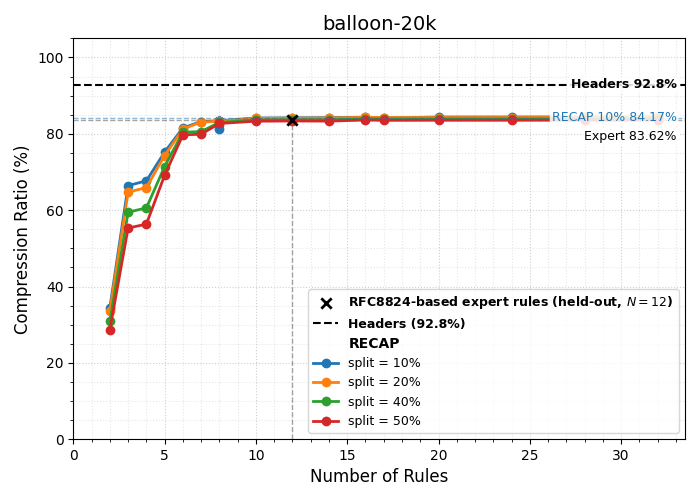}
        \caption{Balloon-20k}
    \end{subfigure}
    \hfill
    \begin{subfigure}[t]{0.49\linewidth}
        \centering
        \includegraphics[width=\linewidth]{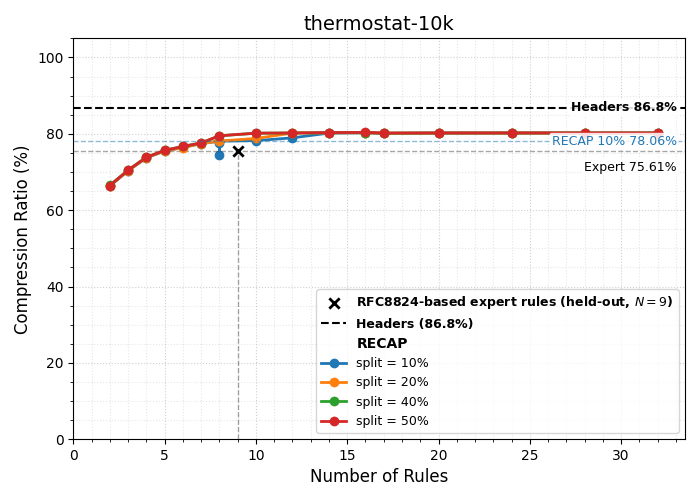}
        \caption{Thermostat-10k}
    \end{subfigure}

    \vspace{0.6em}

    \begin{subfigure}[t]{0.49\linewidth}
        \centering
        \includegraphics[width=\linewidth]{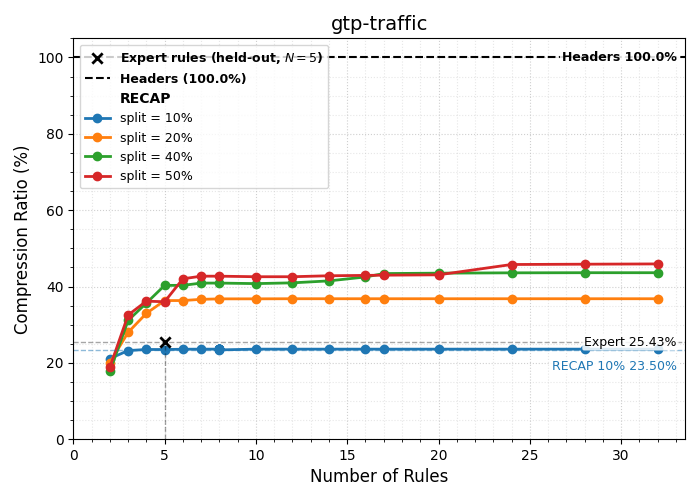}
        \caption{GTP-traffic}
    \end{subfigure}
    \hfill
    \begin{subfigure}[t]{0.49\linewidth}
        \centering
        \includegraphics[width=\linewidth]{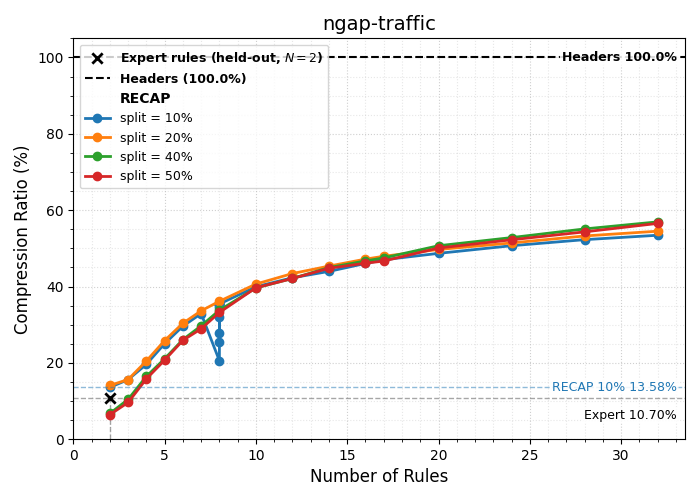}
        \caption{NGAP-traffic}
    \end{subfigure}

\caption{Compression ratio (\%) vs.\ number of rules for each dataset.
        Each line corresponds to a training split ratio and the dashed
        line marks the header ratio. Black crosses mark expert-style
        baselines evaluated on the same held-out test set as RECAP at
        the $10\%$ training split: the RFC~8824 SCHC profile
        for the CoAP datasets ($N^\star{=}12$ for Balloon-20k, $N^\star{=}9$
        for Thermostat-10k), and the transport-header structural baseline
        (IPv4/UDP/GTPv1-fixed for GTP at $N^\star{=}5$, IPv4/SCTP for NGAP
        at $N^\star{=}2$). The head-to-head comparison reads each
        RECAP-curve value for the 10\% split-ratio at the closest grid point
        to $N^\star$. }
    \label{fig:compression-results}
\end{figure}

Figure~\ref{fig:compression-results} reports the compression ratio achieved
on each dataset as a function of the rule budget $N$, for all training splits.
The compression ratio is defined as
$(1 - C_{\text{compressed}} / C_{\text{original}}) \times 100\%$,
where $C_{\text{original}}$ is the total size in bits of the packets
and $C_{\text{compressed}}$ is the total size of the corresponding SCHC-encoded outputs 
(rule identifier plus residue).

\noindent

\paragraph{IoT CoAP datasets.}
On both \textbf{Balloon-20k} and \textbf{Thermostat-10k}, RECAP reaches
within a few percentage points of the header ratio with a moderate number
of rules. On Balloon-20k, $8$ rules (at 10\% split-ratio) achieve $83.4\%$ 
compression on the held-out test set, and $16$ rules reach $84.3\%$ — within $8.6$ 
percentage points of the $92.9\%$ header ratio. Past $N{=}16$ compression saturates
around $84\%$ across splits.

Thermostat-10k is more diverse,
growing from $73.7\%$ at $N=4$ to $80.2\%$ at $N=16$, within $6.6$
percentage points of its $86.8\%$ header ratio. We compare against the expert
baseline obtained by applying RFC~8824~\cite{minaburo_static_2021} --- the
IETF-standardized SCHC profile for CoAP --- field-by-field to each IoT dataset,
with no dataset-specific tuning. This is the rule set a SCHC expert would
deploy by following the standard. RECAP matches or surpasses both expert rule 
sets at the same rule budget. On Balloon-20k, RECAP at the $10\%$ split with 
the expert's $12$-rule budget reaches $84.2\%$ versus $83.6\%$ for the RFC~8824
profile, and continues to climb to $84.3\%$ at $N=16$. On
Thermostat-10k, the expert's $9$-rule budget falls between RECAP
grid points; flanking values are $78.1\%$ at $N=8$ and $78.2\%$ at
$N=10$, both well above the $75.6\%$ baseline, with $80.2\%$
reached at $N=16$.
\paragraph{Core network datasets.}
Unlike CoAP, neither GTP nor NGAP has a standardized SCHC profile to
serve as an expert baseline. We therefore construct a structural
baseline that mirrors what an expert could realistically deploy from
public specifications alone: packets are parsed with a transport-only
parser exposing IPv4/UDP/GTPv1-fixed for GTP and IPv4/SCTP for NGAP,
and one rule is emitted per observed header-descriptor sequence within
that truncated view, with constants elided, recomputable fields marked
for computation, and the remaining transport fields sent verbatim.
Bytes past the transport header (GTP information elements, NGAP/ASN.1
PDUs) are opaque to this baseline, as they would be to an expert
without trace-specific field knowledge. The resulting rule sets are
small ($4$ templates for GTP, $1$ for NGAP) and reach $24.5\%$ and
$10.6\%$ compression on the full traces respectively --- the realistic
floor that any data-driven method must clear to be useful on these
protocols.

The \textbf{GTP-traffic} dataset is small ($100$ packets, so only $10$
training packets at the $10\%$ split): the clusterer identifies one
dominant flow template and achieves $23.5\%$ compression on the held-out
test set, saturating regardless of the rule budget ($N=4$ through $N=32$),
because additional rules cannot describe structure absent from the training
sample. On the matching held-out test set, the structural baseline at
its own budget ($N^\star{=}5$) reaches $25.4\%$, slightly above RECAP
at the $10\%$ split ($23.5\%$ at $N=5$). With only $10$ training
packets, RECAP cannot identify structures that the structural baseline
gets ``for free'' from the published GTP specification. The picture
reverses with more training data: rule sets of size $N=4$ learned from the
$20\%$ split and up reach $33.0\%$ to $36.0\%$ compression, surpassing the 
structural baseline by a growing margin.
The \textbf{NGAP-traffic}
dataset is the most heterogeneous: it carries multiple 5G procedure
types (registration, PDU session, handover, etc.), each with a
different header structure and variable-length information elements.
At the structural baseline's own budget ($N^\star{=}2$), RECAP at the
$10\%$ split already returns $13.6\%$ versus the $10.7\%$ held-out
baseline, and
compression then improves steadily with the rule budget ($13.6\%$ at
$N=2$ to $53.4\%$ at $N=32$ at the $10\%$ split), confirming that
RECAP scales gracefully under rule-budget pressure and dominates the
structural floor by a large margin from the smallest budgets onward.

\paragraph{Hyperparameter sensitivity.}
We default to $\theta{=}0.95$ and $M_{\mathrm{map}}{=}8$ throughout
this section. Appendix~\ref{appendix:sensitivity} sweeps both
parameters at fixed $N{=}8$: held-out ratios are flat in $\theta$
across the range $[0.5, 0.99]$ on the IoT traces and on GTPv1, and
saturate by $M_{\mathrm{map}}{=}8$ on all four traces. NGAP is the
only trace that exhibits visible sensitivity, but the chosen defaults
sit on its saturation plateau as well.
Ablation studies in Appendix~\ref{appendix:ablations} validate the individual RECAP components: our entropy-ratio criterion outperforms raw entropy on structured traces, the dynamic-programming rule selector avoids redundant parent--child selections that plague a greedy alternative, and the Good-Turing coverage estimator is critical for generalizing to held-out traffic on heterogeneous datasets.

\section{Limitations}
\label{sec:limitations}
Our framework is instantiated and evaluated only on SCHC; the two-stage
formulation of Section~\ref{sec:detailed-proposal} extends to any rule-based
compressor consuming structured records, but this is a design observation
rather than an empirical claim. The four-dataset evaluation spans contrasting
operational regimes (IoT telemetry, IoT request/response, training-data
sparsity, heterogeneous 5G signaling) but is a finite sample of the
protocol-traffic space. Performance depends on two hyperparameters, $\theta$
and $M_{\mathrm{map}}$; Appendix~\ref{appendix:sensitivity} shows the chosen
defaults sit on a saturation plateau across all four traces. The dynamic
program runs in $O(|V|\,N^2)$, keeping the full campaign tractable on a
single laptop CPU.

\section{Conclusion}
\label{sec:conclusion}
We presented RECAP, a machine learning framework for automated discovery of SCHC
compression rules. More generally, RECAP learns reusable traffic-pattern
structure that can be translated into rule sets for any compression mechanism
able to exploit such patterns. By combining entropy-guided divisive clustering, a
normalized-entropy stopping criterion, and dynamic programming under a rule budget,
the method learns traffic structure and allocates limited rule memory efficiently.
Across four real-world datasets, RECAP delivers strong compression
with compact rule sets, reducing reliance on manual rule engineering. Future
work includes instantiating the same learning pipeline for non-SCHC
compression frameworks, extending to additional protocol families (e.g.,
QUIC), investigating online or on-device adaptation, and using RECAP-learned
traffic characteristics (especially constant and low-cardinality fields) to
initialize ROHC contexts and profile parameters before online adaptation.

\bibliographystyle{plain}
\bibliography{bibliography.bib}
\newpage
\appendix

\section{Full Per-Dataset Compression Results}
\label{app:full-results}

This appendix tabulates the compression ratios summarized by
Figure~\ref{fig:compression-results}. Table~\ref{table:full-results}
reports, for each dataset and each training split, the held-out
compression ratio (\%) of RECAP across the full rule-budget grid
$N\in\{2,3,4,5,6,7,8,10,12,14,16,17,20,24,28,32\}$. Header ratios
are $92.85\%$ (Balloon-20k), $86.77\%$ (Thermostat-10k), and
$100\%$ for both 5G datasets (parsed packets contain no payload).
The expert-style baselines, evaluated on the same held-out test set
as the $10\%$ split, are $83.62\%$ (Balloon-20k, RFC~8824, $N=12$),
$75.61\%$ (Thermostat-10k, RFC~8824, $N=9$), $25.43\%$ (GTP-traffic,
structural, $N=5$), and $10.70\%$ (NGAP-traffic, structural, $N=2$).

\begin{table}[htbp]
    \centering
    \scriptsize
    \setlength{\tabcolsep}{3.2pt}
    \begin{tabular}{r cccc cccc cccc cccc}
        \toprule
        & \multicolumn{4}{c}{\textbf{Balloon-20k}}
        & \multicolumn{4}{c}{\textbf{Thermostat-10k}}
        & \multicolumn{4}{c}{\textbf{GTP-traffic}}
        & \multicolumn{4}{c}{\textbf{NGAP-traffic}} \\
        \cmidrule(lr){2-5}\cmidrule(lr){6-9}\cmidrule(lr){10-13}\cmidrule(lr){14-17}
        $N$
        & 10\% & 20\% & 40\% & 50\%
        & 10\% & 20\% & 40\% & 50\%
        & 10\% & 20\% & 40\% & 50\%
        & 10\% & 20\% & 40\% & 50\% \\
        \midrule
         2 & 34.4 & 33.5 & 30.8 & 28.7 & 66.3 & 66.4 & 66.5 & 66.4 & 21.1 & 20.1 & 17.8 & 18.8 & 13.6 & 14.1 &  6.7 &  6.4 \\
         3 & 66.4 & 64.7 & 59.5 & 55.3 & 70.3 & 70.3 & 70.5 & 70.5 & 23.2 & 28.0 & 31.1 & 32.5 & 15.6 & 15.6 & 10.4 &  9.6 \\
         4 & 67.7 & 65.9 & 60.6 & 56.3 & 73.7 & 73.7 & 73.9 & 73.8 & 23.5 & 33.0 & 35.7 & 36.2 & 19.6 & 20.5 & 16.5 & 15.8 \\
         5 & 75.2 & 74.3 & 71.4 & 69.1 & 75.5 & 75.5 & 75.7 & 75.6 & 23.5 & 36.3 & 40.3 & 36.0 & 25.0 & 25.8 & 21.1 & 20.7 \\
         6 & 81.5 & 81.3 & 80.4 & 79.7 & 76.4 & 76.4 & 76.7 & 76.7 & 23.6 & 36.3 & 40.3 & 42.0 & 29.6 & 30.4 & 26.1 & 25.9 \\
         7 & 83.2 & 83.1 & 80.6 & 79.9 & 77.4 & 77.4 & 77.6 & 77.6 & 23.6 & 36.6 & 40.9 & 42.7 & 32.9 & 33.6 & 29.7 & 28.9 \\
         8 & 83.4 & 83.3 & 83.0 & 82.7 & 78.1 & 78.1 & 79.5 & 79.5 & 23.6 & 36.8 & 40.9 & 42.7 & 35.4 & 36.1 & 33.8 & 33.2 \\
        10 & 84.1 & 84.0 & 83.6 & 83.3 & 78.2 & 78.8 & 80.1 & 80.2 & 23.6 & 36.8 & 40.8 & 42.6 & 39.9 & 40.6 & 39.6 & 39.7 \\
        12 & 84.2 & 84.1 & 83.7 & 83.4 & 78.9 & 80.2 & 80.2 & 80.2 & 23.6 & 36.8 & 41.0 & 42.6 & 42.3 & 43.4 & 42.1 & 42.1 \\
        14 & 84.2 & 84.2 & 83.7 & 83.3 & 80.2 & 80.2 & 80.3 & 80.3 & 23.6 & 36.8 & 41.5 & 42.8 & 44.0 & 45.3 & 45.0 & 44.8 \\
        16 & 84.3 & 84.3 & 83.9 & 83.6 & 80.2 & 80.3 & 80.3 & 80.4 & 23.6 & 36.8 & 42.5 & 42.9 & 46.0 & 47.2 & 46.7 & 46.1 \\
        17 & 84.1 & 84.3 & 83.9 & 83.5 & 80.1 & 80.1 & 80.2 & 80.2 & 23.6 & 36.8 & 43.4 & 43.0 & 47.0 & 47.9 & 47.5 & 46.7 \\
        20 & 84.4 & 84.3 & 83.9 & 83.5 & 80.2 & 80.2 & 80.2 & 80.2 & 23.6 & 36.8 & 43.5 & 43.1 & 48.7 & 49.7 & 50.7 & 50.1 \\
        24 & 84.4 & 84.3 & 83.9 & 83.5 & 80.2 & 80.2 & 80.2 & 80.2 & 23.6 & 36.8 & 43.6 & 45.8 & 50.7 & 51.4 & 52.8 & 52.2 \\
        28 & 84.4 & 84.3 & 83.9 & 83.6 & 80.2 & 80.2 & 80.2 & 80.2 & 23.6 & 36.8 & 43.6 & 45.8 & 52.3 & 53.2 & 55.1 & 54.3 \\
        32 & 84.4 & 84.3 & 83.9 & 83.6 & 80.2 & 80.2 & 80.2 & 80.2 & 23.6 & 36.8 & 43.6 & 45.9 & 53.4 & 54.5 & 56.9 & 56.5 \\
        \bottomrule
    \end{tabular}
    \caption{RECAP compression ratio (\%) on the held-out test set
    for each dataset, training split, and rule budget $N$.}
    \label{table:full-results}
\end{table}

\section{Dataset Template Hierarchies}
\label{appendix:templates}

This appendix complements the dataset descriptions of Section~\ref{sec:performance-eval}
by exposing the \emph{structural} template hierarchy that RECAP recovers from
each trace, prior to rule selection. Each diagram is the divisive clustering
tree produced on the full trace at entropy ratio
$\theta{=}0.99$ (or $\theta{=}0.95$ and a $2,000$ packets split for NGAP, to keep the figure 
tractable). Direct children of the root are the
\emph{first-level templates}: disjoint groups of packets that share a common
protocol skeleton (a same sequence of fields). Those different structures require
different sets of rules, which an expert RFC-style baseline would identify as
separate ``message types''. Subsequent splits inside each template further identify
sub-structures that can be exploited for compression, and so on recursively.

Because the captures from real cellular core networks are sensitive, all
field-value strings displayed at split points are redacted with the
placeholder ``\guillemotleft\dots\guillemotright''. Field identifiers and
field lengths (in bits) are preserved, as are packet counts on every edge.

\paragraph{Why these four datasets.}
The two IoT traces (\textbf{Balloon-20k}, \textbf{Thermostat-10k}) cover
the canonical SCHC setting: IPv6/UDP/CoAP exchanges with a mix of constant
or low-cardinality fields and unpredictable ones such as
CoAP:Message~ID ($16$~bits, a fresh value per packet), the latter
illustrating why an entropy-\emph{ratio} criterion is required: with $n$
training packets the plug-in estimator yields
$\hat{H} \approx \log_2 n \ll 16$~bits, making such a field look
partially compressible when it carries no redundancy at all.
\textbf{Balloon-20k} is a weather-monitoring scenario dominated by regular
uplink telemetry, representative of LPWAN traffic;
\textbf{Thermostat-10k} mixes CoAP request/response patterns (set-point
configuration, temperature/humidity reporting), exercising rule reuse
under more diverse behavior.
The two 5G datasets move beyond that canonical setting.
\textbf{GTP-traffic} ($100$ packets, PDP-Context management over IPv4/UDP
in a 2G/GPRS core) is an extreme small-sample regime: at the $10\%$
training split only $10$ packets are available, so the underlying
field-value distribution cannot be reliably estimated and the learner
must extract stable, general features from a nearly uninformative sample.
\textbf{NGAP-traffic} ($15{,}650$ packets on the 5G N2 control plane) is
the opposite extreme: many distinct procedure types (registration, PDU
session establishment, handover, NAS transport, etc.), each with its own
header structure and variable-length information elements, so no single
structure dominates and per-structure sample sizes remain small throughout.

\paragraph{Baseline construction.}
On the IoT datasets the baseline is obtained by applying RFC~8824~\cite{minaburo_static_2021} 
recommendations field-by-field. On the 5G
datasets no standardized SCHC profile exists, so we build a rule set from the 
published RFC/3GPP transport-header structure only (IPv4/UDP/GTPv1 fixed header
for GTP, IPv4/SCTP for NGAP). All bytes past these headers --- GTP information
elements, GTP extension headers, NGAP/ASN.1 PDUs --- are treated as
opaque payload, as an expert without trace-specific field knowledge
would see them. For each observed header-descriptor sequence, one rule is created,
with constant fields elided (Equal/Not-Sent), recomputable fields marked Ignore/Compute,
and varying fields Ignore/Value-Sent.

Table~\ref{table:templates} summarizes the key statistics for each dataset, 
including the number of first-level templates.

\begin{table}[htbp]
    \centering
    \begin{tabular}{lccc}
        \toprule
        \textbf{Dataset} & \textbf{Packets used}
            & $\boldsymbol{\theta}$
            & $\boldsymbol{T^{\mathrm{templates}}}$ \\
        \midrule
        Balloon-20k    & $20{,}000$ & $0.99$ & $8$  \\
        Thermostat-10k & $10{,}000$ & $0.99$ & $6$  \\
        GTP-traffic    & $100$      & $0.99$ & $10$ \\
        NGAP-traffic   & $2{,}000$  & $0.95$ & $93$ \\
        \bottomrule
    \end{tabular}
    \caption{First-level structural template counts. Note: NGAP is run on a 
    $2{,}000$-packet subsample (full ASN.1 parsing of the entire $15{,}650$-packet 
    trace yields a hierarchy too large to render legibly); the appendix figure 
    further restricts the rendering to the $20$ most frequent first-level 
    templates and aggregates the remaining $73$ as a single \emph{Others} sink 
    for readability.}
    \label{table:templates}
\end{table}

\paragraph{Balloon-20k.} The hierarchy is shallow and dominated by a small
number of high-volume CoAP uplinks (regular telemetry reports). Splits inside
the dominant template separate request/response pairs and a few sporadic
control exchanges, which is why a budget of $N{=}10$ rules already suffices
to absorb almost all variability (cf. Table~\ref{table:full-results}).

\paragraph{Thermostat-10k.} The trace mixes set-point configuration with
periodic temperature/humidity reporting, so the root expands into more
templates than Balloon-20k at the application layer, but each template is
itself relatively homogeneous. The figure makes visible why a small rule
budget already captures most of the gain: a few dominant templates cover the
vast majority of packets and the remaining structures are rare.

\paragraph{GTP-traffic.} Despite containing only $100$ packets, the trace
exposes $10$ first-level templates corresponding to distinct PDP-Context 
management flows. The tree is unusually wide given the very
small number of packets, which is precisely why the small-sample regime is
challenging: every template carries only a handful of training observations.

\paragraph{NGAP-traffic.} NGAP-traffic is the most heterogeneous of the four
traces: $15{,}650$ packets carrying many distinct 5G control-plane procedures
(registration, PDU session establishment, handover, mobility, NAS transport,
etc.), each with its own header skeleton and variable-length information
elements. On the $2{,}000$-packet subsample used in this appendix
(cf.~Table~\ref{table:templates}), the recovered hierarchy already comprises
$93$ structurally distinct first-level templates, with a
long-tailed packet distribution: the $20$ most frequent templates shown in
the figure account for the bulk of the trace, while the remaining $73$ are
aggregated into a single ``Others'' sink in the figure (though they
participate normally in the clustering). Because no single skeleton dominates,
per-template sample sizes remain small even at this trace size, which is what
makes a small fixed rule budget challenging on NGAP and motivates the
discussion in Section~\ref{sec:performance-eval}.

\begin{figure}[H]
    \centering
    \begin{subfigure}[t]{\textwidth}
        \centering
        \includegraphics[width=\linewidth,height=0.42\textheight,keepaspectratio]{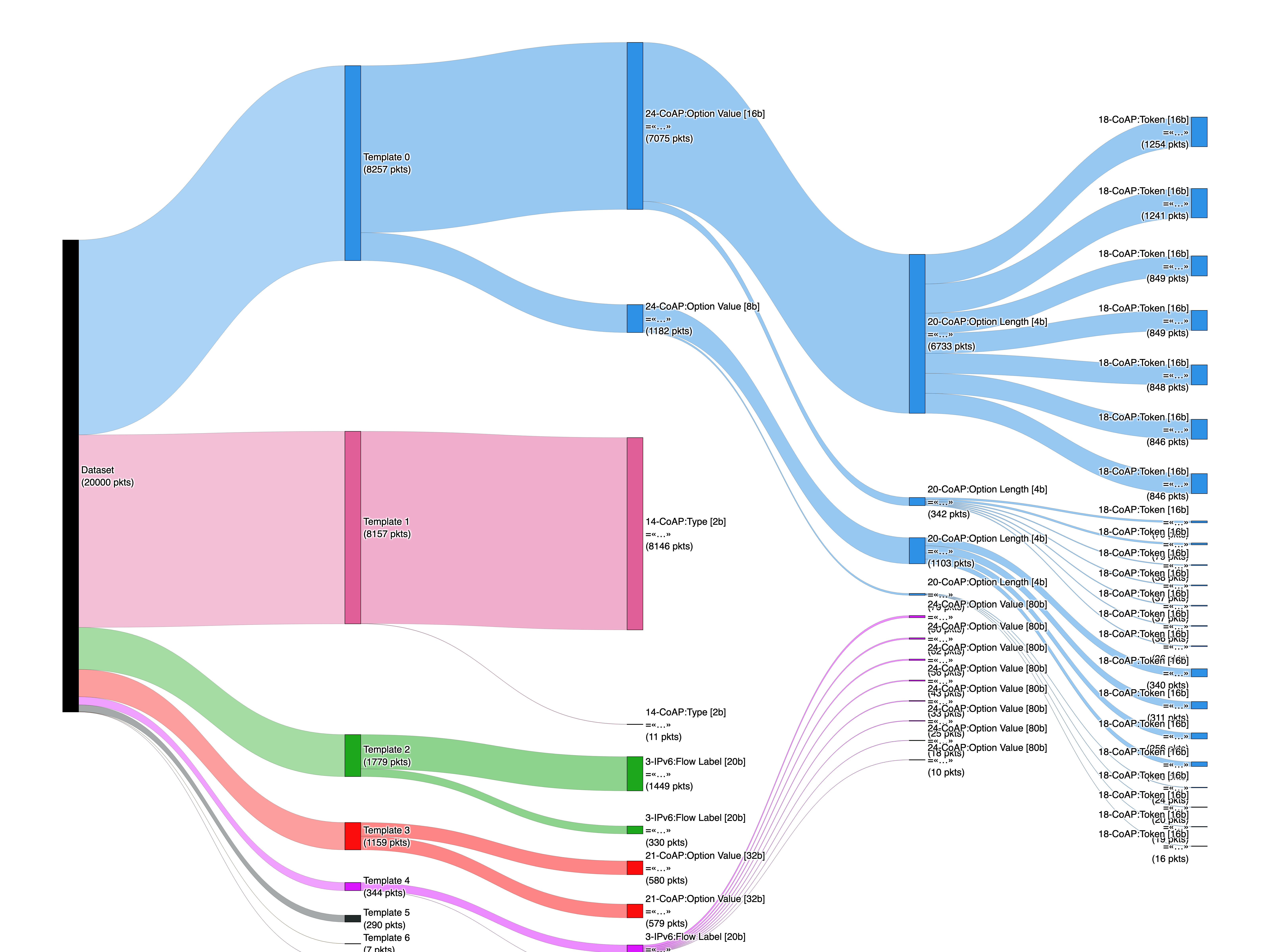}
        \caption{Balloon-20k.}
        \label{fig:sankey-balloon}
    \end{subfigure}

    \vspace{0.8em}
    \begin{subfigure}[t]{\textwidth}
        \centering
        \includegraphics[width=\linewidth,height=0.42\textheight,keepaspectratio]{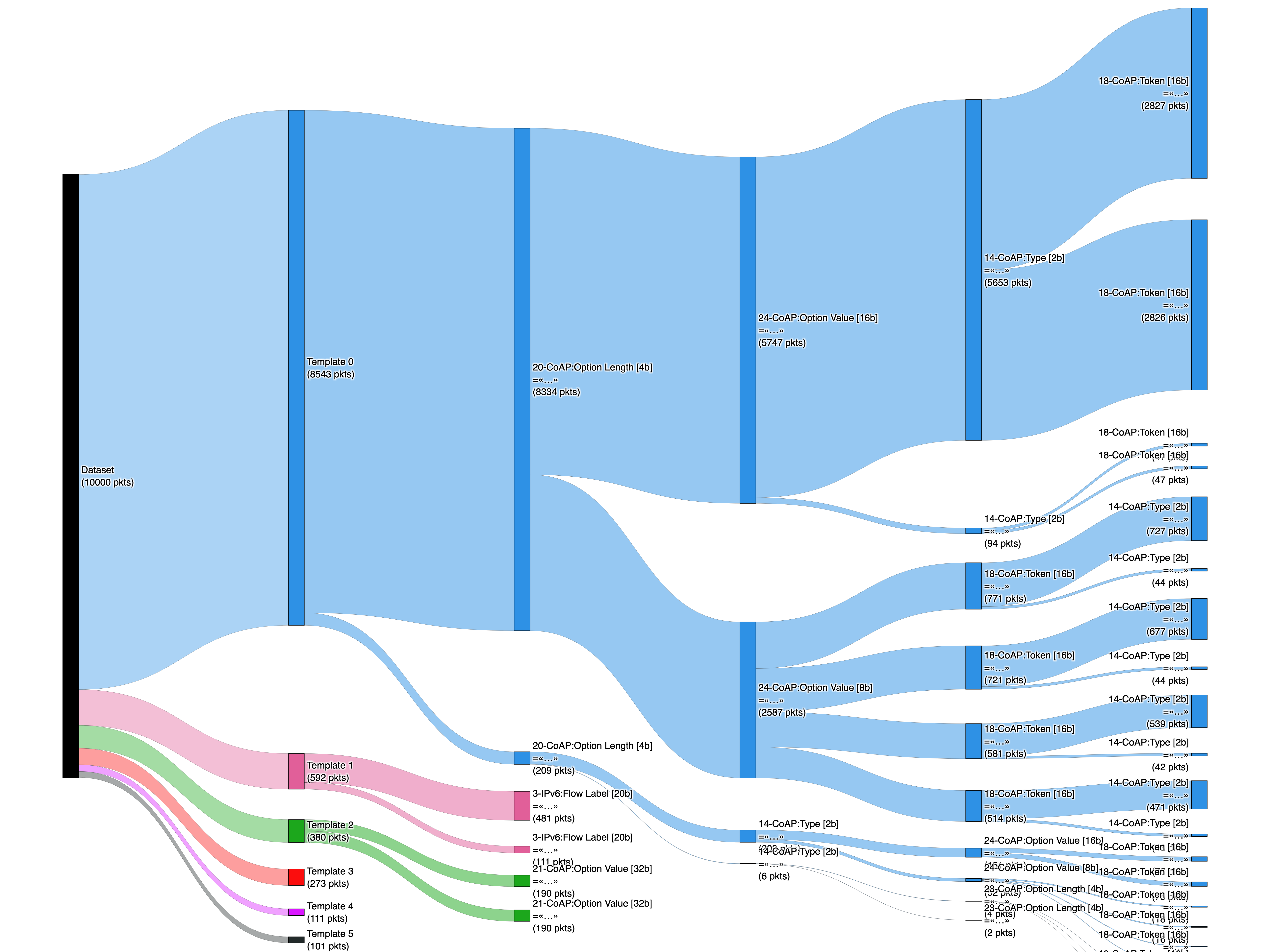}
        \caption{Thermostat-10k.}
        \label{fig:sankey-thermostat}
    \end{subfigure}
    \caption{Structural template hierarchies of the IoT
    datasets. Edges are scaled by packet count; first-level templates are
    direct children of the \emph{Dataset} node and correspond to disjoint
    protocol skeletons. Field-value strings used at split points are
    redacted (\guillemotleft\dots\guillemotright) for legibility; field
    identifiers and lengths (in bits) are preserved.}
    \label{fig:sankey-iot}
\end{figure}

\begin{figure}[H]
    \centering
    \begin{subfigure}[t]{\textwidth}
        \centering
        \includegraphics[width=\linewidth,height=0.40\textheight,keepaspectratio]{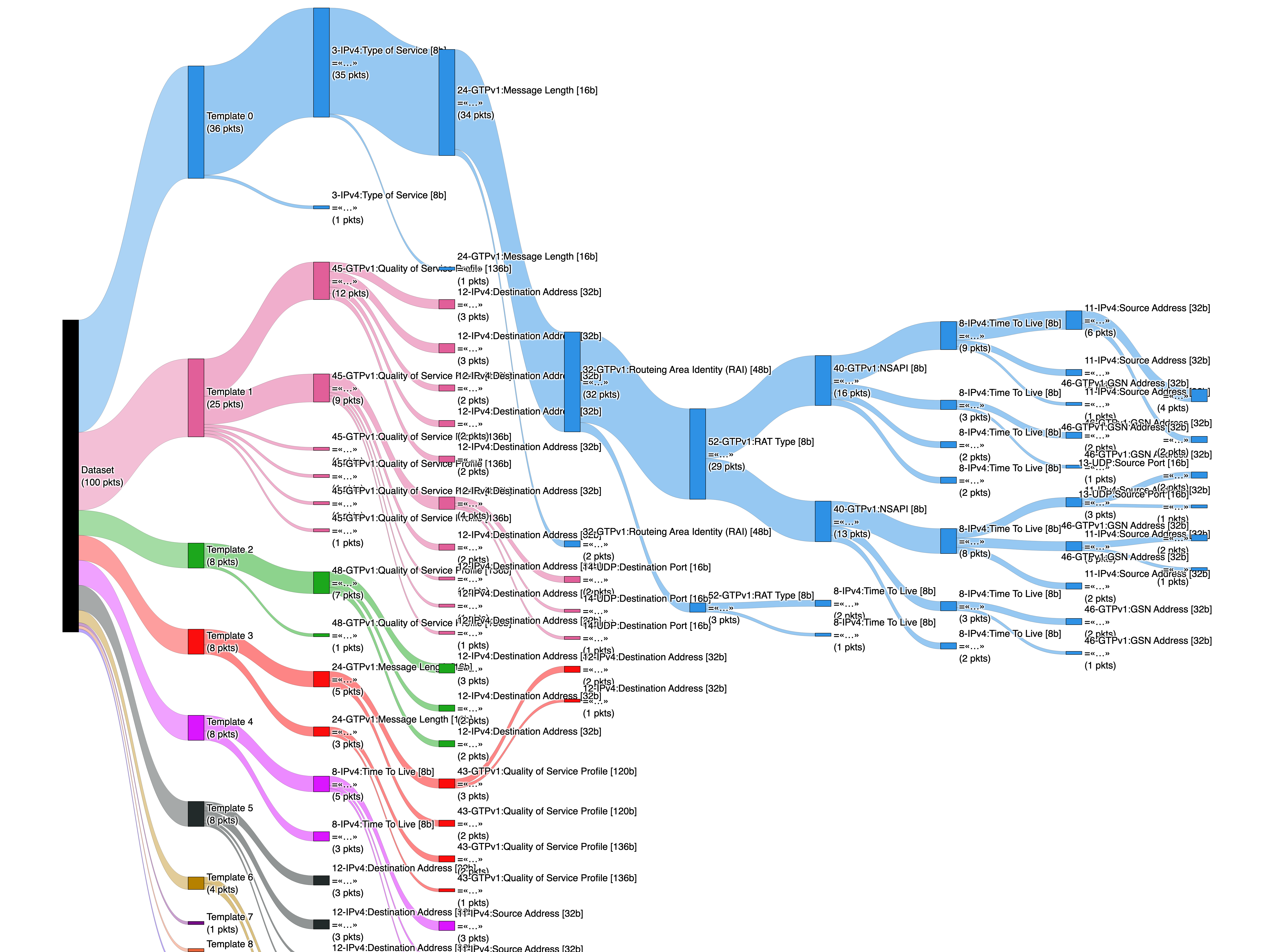}
        \caption{GTP-traffic.}
        \label{fig:sankey-gtp}
    \end{subfigure}

    \vspace{0.8em}
    \begin{subfigure}[t]{\textwidth}
        \centering
        \includegraphics[width=\linewidth,height=0.40\textheight,keepaspectratio]{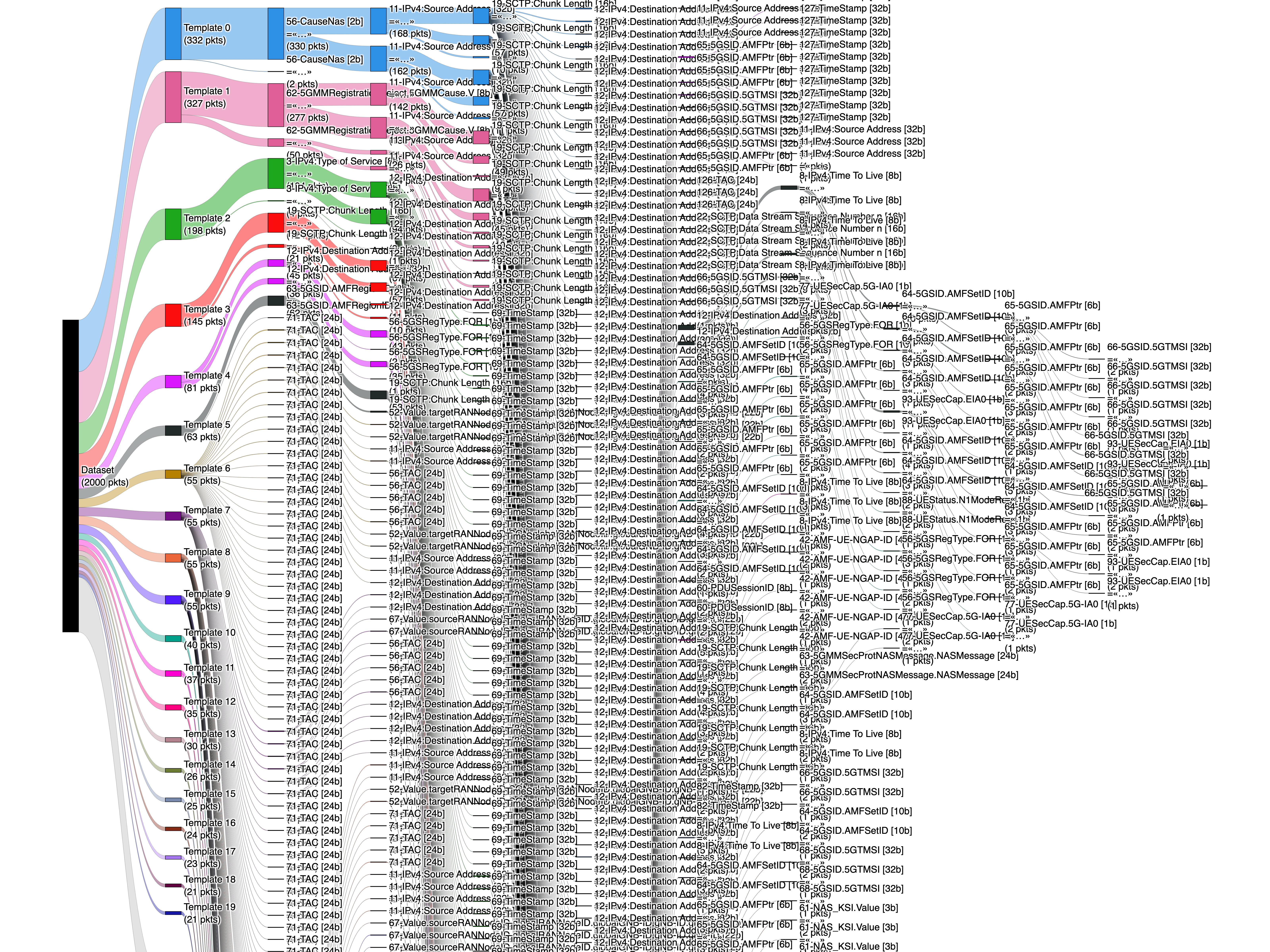}
        \caption{NGAP-traffic (top-$20$ of $93$ first-level templates;
        remaining $73$ aggregated as ``Others'').}
        \label{fig:sankey-ngap}
    \end{subfigure}
    \caption{Structural template hierarchies of the 5G core-network datasets. 
    Edges are scaled by packet count; first-level
    templates are direct children of the \emph{Dataset} node and correspond
    to disjoint protocol skeletons. Field-value strings used at split points
    are redacted (\guillemotleft\dots\guillemotright) for confidentiality;
    field identifiers and lengths (in bits) are preserved.}
    \label{fig:sankey-5g}
\end{figure}

\section{Hyperparameter Sensitivity}
\label{appendix:sensitivity}

This appendix studies the sensitivity of RECAP to its two main
hyperparameters --- the normalized-entropy stopping threshold $\theta$
and the mapping-table cap $M_{\mathrm{map}}$ --- at a fixed rule budget
$N{=}8$ and the $10\%$ training split. All other settings match the
experimental setup of Section~\ref{sec:performance-eval}, and reported
ratios are measured on the held-out test set.

\begin{figure}[htbp]
    \centering
    \begin{subfigure}[t]{0.49\linewidth}
        \centering
        \includegraphics[width=\linewidth]{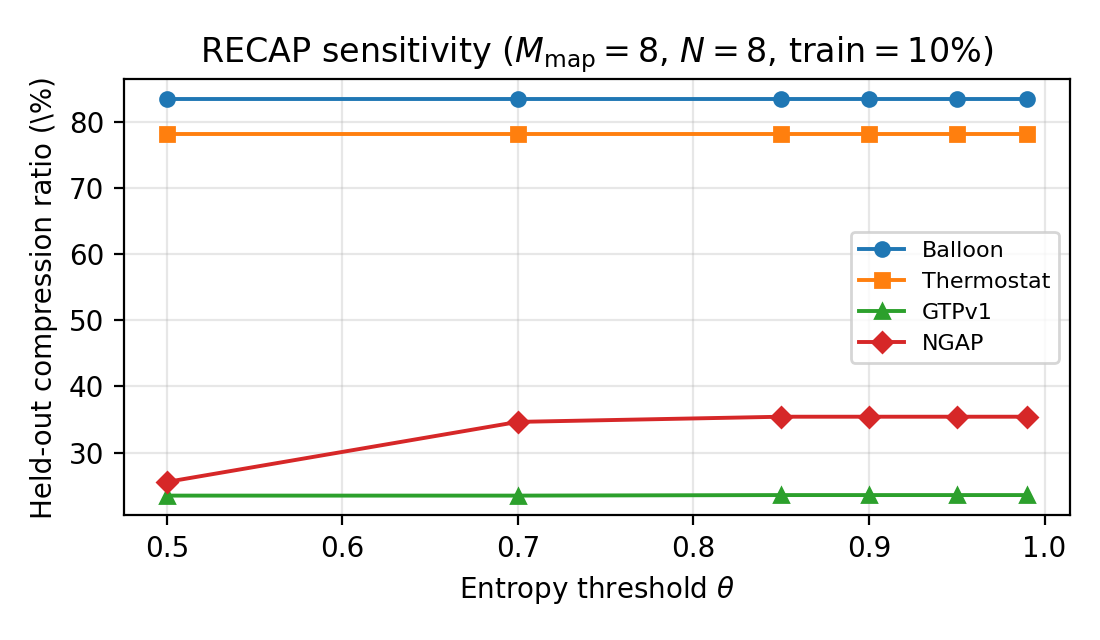}
        \caption{Sensitivity to the entropy threshold $\theta$ at fixed
        $M_{\mathrm{map}}{=}8$. Only NGAP shows visible sensitivity, and
        only below $\theta{=}0.85$.}
        \label{fig:sensitivity-theta}
    \end{subfigure}
    \hfill
    \begin{subfigure}[t]{0.49\linewidth}
        \centering
        \includegraphics[width=\linewidth]{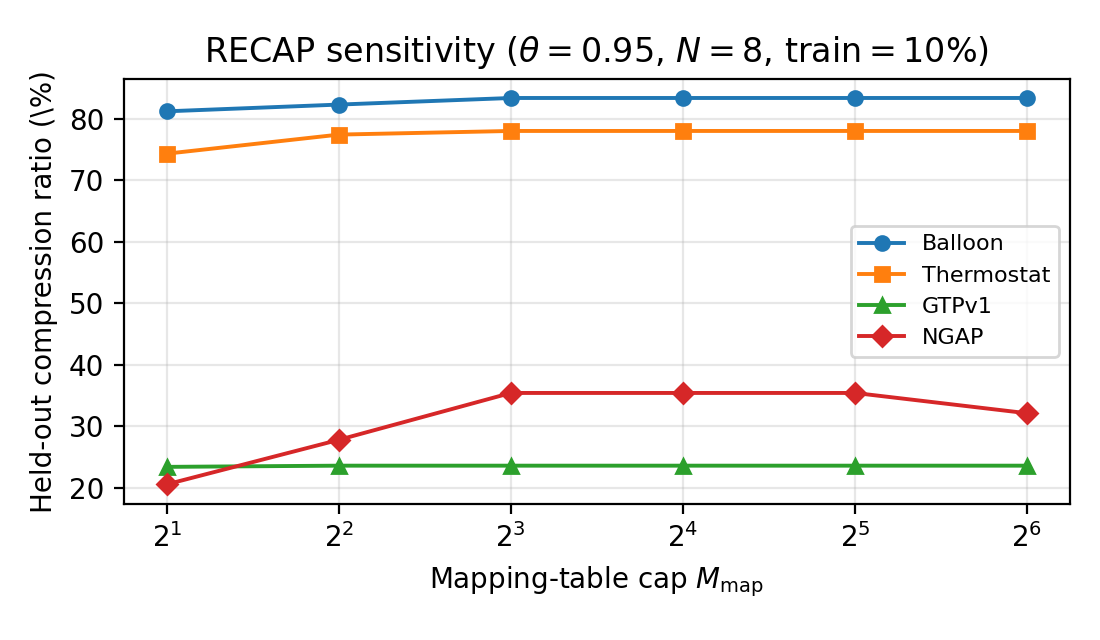}
        \caption{Sensitivity to the mapping-table cap $M_{\mathrm{map}}$
        at fixed $\theta{=}0.95$. Performance saturates by
        $M_{\mathrm{map}}{=}8$ on all traces; NGAP-Traffic is the only trace
        that pays a substantial penalty at smaller caps.}
        \label{fig:sensitivity-mapping}
    \end{subfigure}
    \caption{RECAP held-out compression ratio (\%) versus each
    hyperparameter, at fixed $N{=}8$ and $10\%$ training split.}
    \label{fig:sensitivity}
\end{figure}

\paragraph{Stopping threshold $\theta$.}
Figure~\ref{fig:sensitivity-theta} reports the held-out compression ratio
as $\theta$ varies in $\{0.5, 0.7, 0.85, 0.9, 0.95, 0.99\}$. Lower values
prune candidate splits aggressively (only fields with very low
normalized entropy are split), while higher values retain more candidates
and defer the trade-off to the rule-selection stage. On the two IoT
traces and on GTP-Traffic the curves are essentially flat: at $N{=}8$ the
dynamic-programming selector picks the same effective partition
regardless of which low-value splits remain available. NGAP-traffic is the
only trace where $\theta$ matters in this range: aggressive pruning
at $\theta{=}0.5$ drops the held-out ratio to $25.5\%$, whereas any
value $\theta{\geq}0.85$ recovers the plateau near $35.4\%$. The
choice $\theta{=}0.95$ used in the main experiments lies on this
plateau across all four traces. We do not include $\theta{=}1.0$ in
the sweep: it disables pruning and makes the candidate hierarchy
explode, which is the regime the threshold is designed to avoid.

\paragraph{Mapping-table cap $M_{\mathrm{map}}$.}
Figure~\ref{fig:sensitivity-mapping} reports the same metric as
$M_{\mathrm{map}}$ varies in $\{2,4,8,16,32,64\}$ at $\theta{=}0.95$.
Raising the cap allows \texttt{Match-Mapping}/\texttt{Mapping-Sent} to
absorb fields with more distinct values at the cost of a slightly
longer mapping index inside the compressed header. The IoT traces and
GTP-Traffic saturate by $M_{\mathrm{map}}{=}8$; only NGAP-Traffic, which has by far
the largest template diversity (cf.\ Table~\ref{table:templates}),
visibly benefits from values $M_{\mathrm{map}}{\geq}8$ and pays a
substantial penalty at the smallest caps ($20.6\%$ at
$M_{\mathrm{map}}{=}2$ versus $35.4\%$ at $M_{\mathrm{map}}{=}8$).
Beyond $M_{\mathrm{map}}{=}8$ none of the traces gain measurably,
confirming that the value used in the main experiments is a sound
default.

\section{Comparison with General-Purpose Compressors}
\label{appendix:general-compression}

The learning problem addressed in this paper is specific to rule-based
lossless compression, instantiated here as SCHC rules. To our knowledge,
there is currently no learning baseline for this specific problem.
General-purpose compression mechanisms (e.g., GZip, Zlib) operate on
fundamentally different assumptions, such as compressing large files where
the cost of building and storing a dictionary alongside the residue is
amortized. These assumptions hinder their applicability for network traffic,
where packet sizes are small and context must be shared. Nonetheless, we
provide GZip and Zlib results below, as they substantiate our claims
regarding the regime difference.

\begin{table}[htbp]
\centering
\caption{Per-packet header-only compression comparison. Each packet is compressed independently (self-contained, no shared state). Negative values indicate expansion (compressed output larger than input).}
\label{table:general-compression}
\begin{tabular}{@{}lcccccc@{}}
\toprule
\textbf{Dataset} & \textbf{Avg HDR} & \textbf{Gzip\textsubscript{HDR}} &
\textbf{Gzip\textsubscript{FULL}} & \textbf{Zlib\textsubscript{HDR}} &
\textbf{Zlib\textsubscript{FULL}} & \textbf{RECAP} ($N{=}32$) \\
\midrule
Balloon-20k      & 57.2B   & $-1.7\%$   & $-0.6\%$   & $19.2\%$ & $18.8\%$ & \textbf{84.4\%} \\
Thermostat-10k   & 60.4B   & $-1.7\%$   & $0.9\%$   & $18.2\%$ & $18.2\%$ & \textbf{80.2\%} \\
GTP-traffic      & 97.3B   & $-13.4\%$  & $-13.4\%$  & $-1.1\%$ & $-1.1\%$ & \textbf{23.6\%} \\
NGAP-traffic     & 173.2B  & $6.1\%$   & $6.1\%$   & $13.1\%$& $13.1\%$& \textbf{53.4\%} \\
\bottomrule
\end{tabular}
\end{table}

Each packet is compressed independently (self-contained, no shared state).
General-purpose compressors rely on per-packet dictionaries and sliding
windows to find redundancies. Because the output must be decompressible on
its own, the compressor embeds its internal state (e.g., Huffman tables,
length/distance codes, LZ77 window state) alongside the compressed payload.
This structural overhead dominates \textasciitilde50\,--\,170B headers,
causing negative compression (packet expansion) on most datasets.
RECAP dominates by a wide margin, confirming that SCHC's rule-based
approach --- amortizing a shared context across thousands of packets --- is
fundamentally better suited to header compression than per-packet statistical methods.
\section{Ablation Studies}
\label{appendix:ablations}

This appendix validates the individual components of the RECAP pipeline
by replacing each one with a simpler alternative and measuring the
impact on held-out compression ratio at a fixed rule budget and
$10\%$ training split.

\paragraph{Greedy Top-$k$ vs.\ Dynamic Programming.}
Our DP rule selector avoids redundant parent--child selections: it
optimizes the joint allocation of the budget across the candidate
tree. Greedy Top-$k$ evaluates each candidate independently by its
standalone gain and picks the $k$ highest-scoring nodes. On shallow,
homogeneous trees the two strategies are close; the DP advantage emerges
when the candidate hierarchy is deep and heterogeneous, because Greedy
can waste budget slots on a parent cluster whose gain overlaps with
its children.

\begin{table}[htbp]
\centering
\caption{DP vs.\ Greedy Top-$k$ rule selection ($N{=}8$, $\theta{=}0.95$, $10\%$ split).}
\label{table:dp-vs-greedy}
\begin{tabular}{@{}lccccc@{}}
\toprule
\textbf{Dataset} & \textbf{DP} & \textbf{Greedy} & \textbf{Gap} &
\textbf{DP Rules} & \textbf{Greedy Rules} \\
\midrule
Balloon-20k    & $83.42\%$ & $83.24\%$ & $+0.18$ pp & 7 & 7 \\
Thermostat-10k & $78.06\%$ & $78.06\%$ & $0.00$  pp & 7 & 7 \\
GTP-traffic    & $23.59\%$ & $23.59\%$ & $0.00$  pp & 6 & 7 \\
NGAP-traffic   & $35.42\%$ & $35.42\%$ & $0.00$  pp & 7 & 7 \\
\bottomrule
\end{tabular}
\end{table}

At $N{=}8$ the gap is $\le 0.18$ pp because the top structural templates
capture most of the gain. The DP advantage grows at larger budgets where
deeper structural splits become available and the joint allocation matters.

\paragraph{Normalized Entropy-Ratio vs.\ Raw Entropy.}
Raw entropy $\hat{H}(X_j)$ is biased by field width: it may prefer a 1-bit
flag with two equally likely values ($H{=}1.0$) over a 128-bit address field
with four values ($H{=}2.0$), even though the address field is far more
compressible per bit. Our entropy-ratio criterion normalizes by
$\min(\hat{L}_j, \log_2(n_c))$ to identify the truly compressible split.

\begin{table}[htbp]
\centering
\caption{Normalized entropy-ratio vs.\ raw entropy splitting criterion ($10\%$ split).}
\label{table:entropy-ablation}
\begin{tabular}{@{}lcccc@{}}
\toprule
\multicolumn{1}{l}{} & \multicolumn{2}{c}{\textbf{Balloon-20k}} & \multicolumn{2}{c}{\textbf{Thermostat-10k}} \\
\cmidrule(lr){2-3}\cmidrule(lr){4-5}
\textbf{$N$} & RECAP & Raw & RECAP & Raw \\
\midrule
8  & $83.42\%$ & $83.42\%$ & $78.12\%$ & $79.40\%$ \\
16 & $84.20\%$ & $83.20\%$ & $80.20\%$ & $80.10\%$ \\
32 & $84.43\%$ & $83.33\%$ & $80.20\%$ & $80.00\%$ \\
\bottomrule
\end{tabular}
\end{table}

On Balloon, the gap opens with the budget: at $N{=}32$ RECAP leads by
$1.1$ pp because its deeper tree (29 leaves) supplies richer candidates
than the raw-entropy tree (8 leaves at $\theta{=}0.95$). On Thermostat,
raw entropy is slightly ahead at $N{=}8$ but the strategies converge at
$N{\ge}16$. On GTP and NGAP the two criteria produce identical compression
across all budgets: the structural templates dominate, and the entropy
criterion mainly refines within them.

\paragraph{Structural Naive Baseline.}
A naive structural baseline groups training packets by template signature
(field-descriptor sequence), selects the $N$ most frequent templates,
and generates full rules via the standard rule generator
(including Match-Mapping for low-cardinality fields). This baseline has
no divisive clustering: each template group is treated as a monolithic
cluster. The gap measures the value of entropy-driven recursive splitting
and the DP budget optimizer over simple template selection.

\begin{table}[htbp]
\centering
\caption{Structural template baseline vs.\ RECAP ($N{=}8$, $10\%$ split).}
\label{table:naive-baseline}
\begin{tabular}{@{}lccc@{}}
\toprule
\textbf{Dataset} & \textbf{Structural} & \textbf{RECAP} & \textbf{Gap} \\
\midrule
Balloon-20k    & $82.19\%$ & $83.42\%$ & $+1.23$ pp \\
Thermostat-10k & $77.43\%$ & $78.06\%$ & $+0.63$ pp \\
GTP-traffic    & $23.51\%$ & $23.59\%$ & $+0.08$ pp \\
NGAP-traffic   & $37.86\%$ & $37.86\%$ & $0.00$  pp \\
\bottomrule
\end{tabular}
\end{table}

The gap is largest on IoT traces where entropy-driven splitting discovers
structure within templates (e.g., CoAP option patterns) that pure
template selection misses. On 5G traces (GTP, NGAP) the structural
templates already capture the compressible signal and the gap vanishes.
\paragraph{Banerjee et al.\ Flat Clustering Baseline.}
Banerjee et al.~\cite{banerjee_automated_2024} propose a flat clustering approach: packets are pre-grouped by template signature, then clustered within each template using Gower field-by-field distances and K-Means. One SCHC rule is constructed per cluster via a heuristic. We adapted this approach to our framework with a fixed rule budget $N$ and $10\%$ training split.

\begin{table}[htbp]
\centering
\caption{Banerjee et al.\ flat clustering vs.\ RECAP ($10\%$ split).}
\label{table:banerjee-baseline}
\begin{tabular}{@{}lcccc@{}}
\toprule
\multicolumn{1}{l}{} & \multicolumn{2}{c}{\textbf{Balloon-20k}} & \multicolumn{2}{c}{\textbf{GTP-traffic}} \\
\cmidrule(lr){2-3}\cmidrule(lr){4-5}
\textbf{$N$} & RECAP & Banerjee & RECAP & Banerjee \\
\midrule
2  & $34.42\%$ & $34.42\%$ & $21.11\%$ & $21.11\%$ \\
4  & $67.67\%$ & $74.18\%$ & $23.51\%$ & $23.51\%$ \\
8  & $83.42\%$ & $83.42\%$ & $23.59\%$ & $6.25\%$ \\
16 & $84.34\%$ & $84.34\%$ & $23.59\%$ & $4.63\%$ \\
32 & $84.43\%$ & $84.17\%$ & $23.59\%$ & $4.63\%$ \\
\bottomrule
\end{tabular}
\begin{tabular}{@{}lcccc@{}}
\toprule
\multicolumn{1}{l}{} & \multicolumn{2}{c}{\textbf{Thermostat-10k}} & \multicolumn{2}{c}{\textbf{NGAP-traffic}} \\
\cmidrule(lr){2-3}\cmidrule(lr){4-5}
\textbf{$N$} & RECAP & Banerjee & RECAP & Banerjee \\
\midrule
2  & $66.32\%$ & $66.32\%$ & $13.58\%$ & $-0.07\%$ \\
4  & $73.70\%$ & $73.70\%$ & $19.62\%$ & $-0.15\%$ \\
8  & $78.06\%$ & $78.00\%$ & $35.42\%$ & $-0.22\%$ \\
16 & $80.24\%$ & $80.06\%$ & $46.03\%$ & $-0.29\%$ \\
32 & $80.20\%$ & $80.08\%$ & $53.42\%$ & $-0.36\%$ \\
\bottomrule
\end{tabular}
\end{table}

On homogeneous IoT traces (Balloon, Thermostat) flat clustering is competitive with RECAP ($\le 0.25$ pp gap at $N{=}32$), because few structural templates exist and K-Means within each template captures the remaining variance. On heterogeneous 5G traces the approach collapses: on GTP, the budget fragments across the $10$ training packets and generalization vanishes at $N{\ge}8$ ($+17.34$ pp gap); on NGAP, the $90$ templates consume the entire budget with no intra-template splitting, producing \emph{negative} compression as most test packets fall through to no-compression. RECAP's recursive, entropy-driven splitting avoids this fragmentation by building a candidate tree first and then optimizing the budget allocation across it.
\paragraph{Good-Turing Coverage vs.\ Uniform Coverage.}
Replacing the Good-Turing coverage estimator $\hat{C}(u)$ with a
uniform estimate ($\hat{C}(u){=}1.0$ for all clusters) removes the
mechanism that down-weights sparse, poorly-generalizing rules. On IoT
traces with large training sets, coverage is near $1.0$ for all clusters
and the difference is negligible. On heterogeneous 5G traces, uniform
coverage is catastrophic: the DP selects rules that match many training
packets but fire rarely on held-out traffic.

\begin{table}[htbp]
\centering
\caption{Good-Turing vs.\ uniform coverage estimator ($N{=}8$, $\theta{=}0.95$, $10\%$ split).}
\label{table:good-turing-ablation}
\begin{tabular}{@{}lccc@{}}
\toprule
\textbf{Dataset} & \textbf{Good-Turing} & \textbf{Uniform} & \textbf{Gap} \\
\midrule
Balloon-20k    & $83.42\%$ & $83.42\%$ & $0.00$  pp \\
Thermostat-10k & $78.06\%$ & $78.15\%$ & $-0.09$ pp \\
GTP-traffic    & $23.59\%$ & $2.27\%$  & $+21.32$ pp \\
NGAP-traffic   & $35.42\%$ & $31.04\%$ & $+4.38$  pp \\
\bottomrule
\end{tabular}
\end{table}

On GTP, uniform coverage collapses compression from $23.59\%$ to
$2.27\%$: the DP selects rules that look strong on the 10-packet
training set but never fire on the 90-packet test set. This validates
the Good-Turing design choice: conservative coverage estimation is
essential for generalizing beyond the training distribution.

\end{document}